\documentclass{article}
\usepackage[utf8]{inputenc}
\usepackage[margin=1in]{geometry}
\usepackage{listings}
\usepackage{graphicx}
\graphicspath{ {images/} }
\usepackage{url}
\usepackage{multicol}
\usepackage{makecell}
\usepackage{subcaption}
\usepackage{float}
\usepackage{hyperref}
\usepackage{authblk}
\usepackage{lipsum}

\usepackage[USenglish]{babel}    
\usepackage{csquotes}
\usepackage[backend=biber]{biblatex}
\bibliography{references}

\begin{document}
\title{Working Paper: NLP in Human Rights Research - Extracting Knowledge Graphs About Police and Army Units and Their Commanders}

\author[1]{Bauer, D.\thanks{Equal Contribution, bauer@cs.columbia.edu}}
\author[2]{Longley, T.\thanks{Equal Contribution, tom@securityforcemonitor.org}}
\author[1]{Ma, Y.\thanks{Equal Contribution, ym2745@columbia.edu}}
\author[2]{Wilson T.\thanks{Equal Contribution, tony@securityforcemonitor.org}}
\affil[1]{Department of Computer Science, Columbia University}
\affil[2]{Security Force Monitor, Human Rights Institute, Columbia Law School}

\date{January 2022}

\maketitle

\let\oldabstract\abstract
\let\oldendabstract\endabstract
\makeatletter
\renewenvironment{abstract}
{\renewenvironment{quotation}%
               {\list{}{\addtolength{\leftmargin}{7em} 
                        \listparindent 1.5em%
                        \itemindent    \listparindent%
                        \rightmargin   \leftmargin%
                        \parsep        \z@ \@plus\p@}%
                \item\relax}%
               {\endlist}%
\oldabstract}
{\oldendabstract}
\makeatother

\begin{abstract}
In this working paper we explore the use of an NLP system to assist the work of Security Force Monitor (SFM).  SFM creates data about the organizational structure, command personnel and operations of police, army and other security forces, which assists human rights researchers, journalists and litigators in their work to help identify and bring to account specific units and personnel alleged to have committed abuses of human rights and international criminal law.  This working paper presents an NLP system that extracts from English language news reports the names of security force units and the biographical details of their personnel, and infers the formal relationship between them.  Published alongside this working paper are the system's code and training dataset.  We find that the experimental NLP system performs the task at a fair to good level. Its performance is sufficient to justify further development into a live workflow that will give insight into whether its performance translates into savings in time and resource that would make it an effective technical intervention.
\end{abstract}

\newpage

\begin{multicols}{2}

\section{Introduction}
Human rights organizations around the world gather large amounts of information for the purposes of promoting and protecting human rights. The promise offered by automated information extraction and processing technologies is of making these rivers of information  easier to comprehend and take action on. This promise is much touted; it also feels like such capacities might be more accessible to everyone in a world where software can drive cars or defeat a 9 \textit{dan} ranked Go master!  What, however, does this promise mean in practice for the basic daily work of human rights researchers, rather than their counterparts in  commercial, scientific and industrial domains?  In this working paper we try to provide some insight into this question by reporting the initial outcomes of a multi-disciplinary collaboration to explore the value of Natural Language Processing (NLP) methods as components of information extraction systems used gather detailed data about state security and defense forces implicated in human rights abuses. 

\begin{figure}[H]
\centering
\includegraphics[width=8cm]{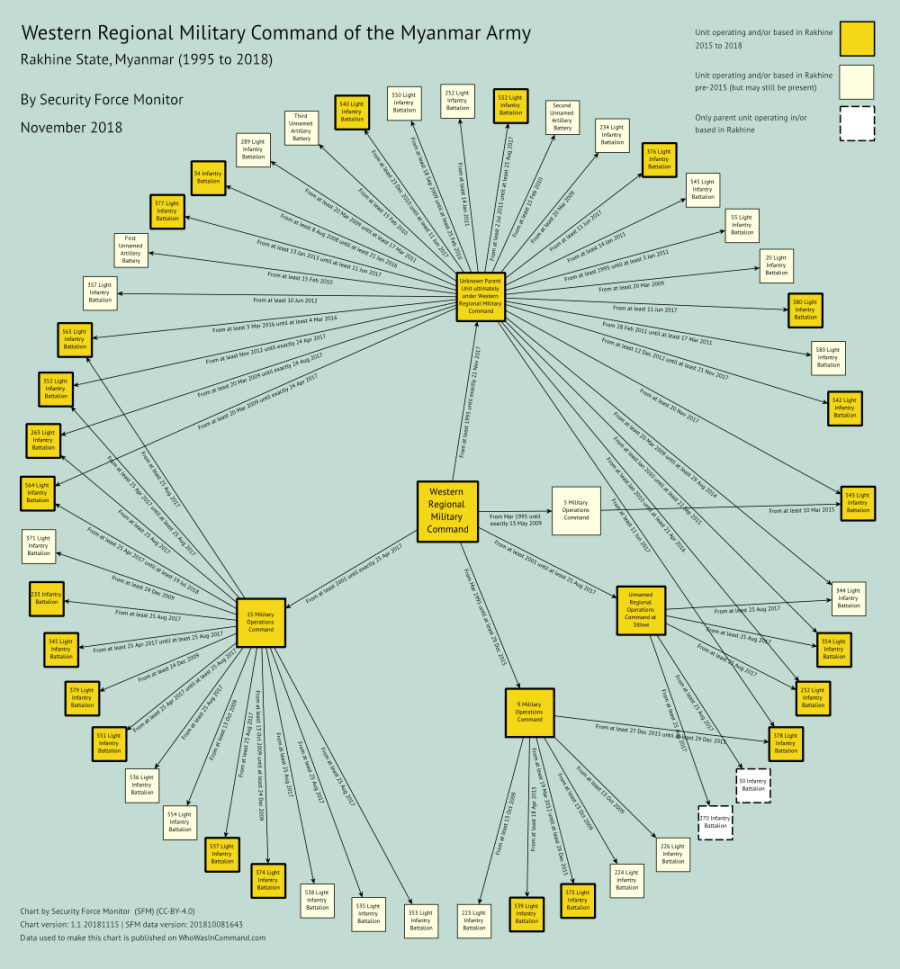}
\caption{Western Regional Military Command of the Myanmar Army – Rakhine State (1995 to 2018) – a chart by Security Force Monitor (CC-BY-4.0) \cite{myanmar2018}}
\label{fig:sfm_my_chart}
\end{figure}

Security Force Monitor\footnote{Security Force Monitor is part of the Columbia Law School Human Rights Institute (\url{https://securityforcemonitor.org}).} (SFM) \cite{wilson_2017} is a human rights project that compiles and analyzes public information to create detailed data on the organizational structure, command personnel and operations of police, army and other security forces. They provide this data to other human rights researchers, investigative journalists and litigators to help them identify and bring to account specific units and personnel alleged to have committed abuses of human rights and international criminal law.  SFM's research has been used in the investigation of drug-related killings by police in the Philippines \cite{philippines2019}, allegations of war crimes committed by the army in Mexico \cite{mexico2018}, and the use of lethal force by the Nigerian military against protesters in Nigeria \cite{nyt2018}.

\begin{figure}[H]
\centering
\includegraphics[width=8cm]{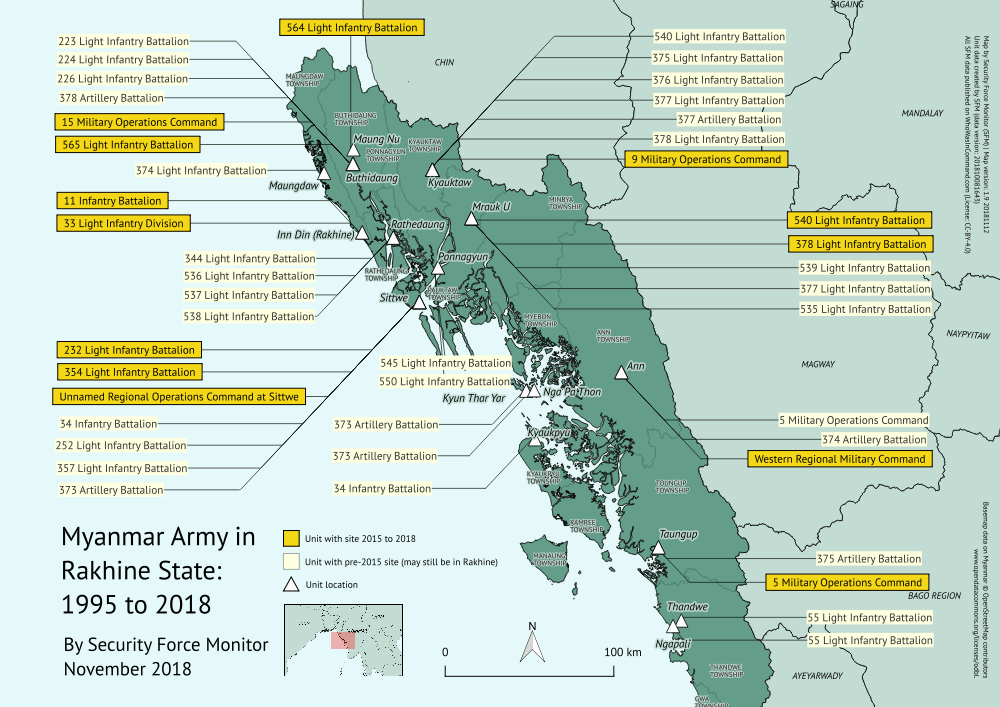}
\caption{Western Regional Military Command of the Myanmar Army – Rakhine State (1995 to 2018) – a map by Security Force Monitor (CC-BY-4.0) \cite{myanmar2018}}
\label{fig:sfm_my_map}
\end{figure}

SFM's approach\footnote{See ``Research Handbook for Security Force Monitor", \url{https://help.securityforcemonitor.org/}}
is to identify salient material (``sources'') through targeted web searches, extract up to 80 specific pieces of information about people, locations and organizations from these sources, and arrange them into a graph-like data structure. These data are transformed into hierarchical organograms or other visualizations of security force structures, showing additional information about personnel (name, rank, role, title), geographic footprint (facilities, bases, camps) and areas of operation. For example, figures \ref{fig:sfm_my_chart} and \ref{fig:sfm_my_map} show the command structure and areas of operation of the Western Regional Military Command of the Myanmar Army.

SFM performs this extraction work mostly ``by hand''. For example, in the course of its research SFM would review the following extract from an article published in Nigeria's \textit{Vanguard} newspaper in 2012 \cite{vanguard2012}: 

\begin{quote}
General Officer Commanding 3 Armoured Division of the Nigerian Army, Major General Jack Nwaogbo, has again re-assured Nigerians that the Boko Haram insurgency would soon be contained.
\end{quote}

From this, SFM  would extract the following pieces of information and enter them into a database:
\begin{itemize}
    \setlength\itemsep{-0.5em}
    \item Name of person: ``Jack Nwaogbo''
    \item Rank of person: ``Major General''
    \item Title of person: ``General Officer Commanding''
    \item Organization/unit: ``3 Armoured Division''
    \item Role of person in unit: ``Commander''
\end{itemize}

SFM also applies a number of integrity measures to every data point: they are specifically evidenced by one or more sources, are time-bound (valid from, valid until) and are rated for confidence in their accuracy (from low to high). SFM also extracts and encodes geographical information about the emplacements and operations of specific units. The resulting datasets are made public by SFM, and can be searched through a public website\footnote{\url{https://WhoWasInCommand.com }}. At the time of writing, SFM has manually analyzed over 8,000 documents, assembling data on 10,900 specific units, 2,700 command personnel and over 200 alleged human rights violations in 19 countries, going back a decade.  Already faced with managing a rich and complicated dataset, SFM faces challenges of scale on numerous fronts: extending coverage to include new force branches and new countries, updating existing data as relevant new material appears online, and working in a number of different languages.

Given the centrality of this type of text analysis to SFM's research process, NLP would seem to hold potential in automating - partly, or fully - time-consuming tasks like identifying and relating a specific person to a specific unit, and extracting contextual biographical data such as rank and official title. The value to SFM is in picking out common named entities (like Persons and Organizations), and in establishing and extracting the relationships between them in a format that can be quickly appraised for accuracy.

This working paper explores this potential in the form of a pilot study and is organized as follows. Section \ref{section:dataset} describes the research task and an annotated dataset of 130 news reports about the defense and security forces of Nigeria, which is released with this paper.  Section \ref{section:pilotstudy} presents initial results of a pilot/baseline system implemented by the authors. We show results system's with respect to both named entity recognition and relation extraction tasks. Finally, in Section \ref{section:expresults} we look at some of the limitations of the system and the experimental results.

\section{Data and Task Description}
\label{section:dataset}

\begin{figure}[H]
\centering
\includegraphics[width=8cm]{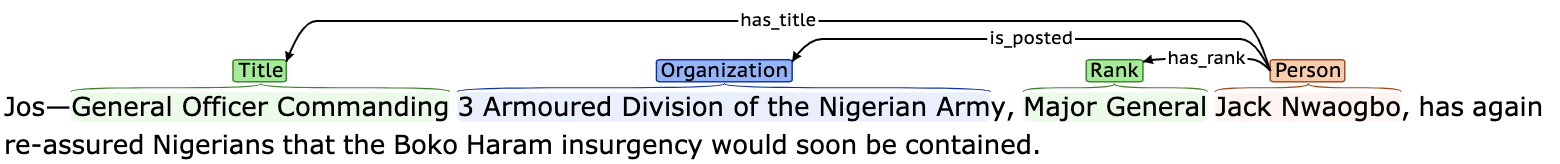}
\caption{An example from our dataset}
\label{fig:brat_vis_screenshot}
\end{figure}

SFM identified 130 of the most information-rich text documents from which it has extracted material to create its data on the Nigerian Army and Nigerian Police Force. They annotated the text of the articles to create a gold standard dataset for use in the development of an NLP system. The annotations describe the relationships between persons and units, which is one of the main information extraction tasks done by hand by SFM. SFM have published the corpus of annotated texts online.\footnote{\url{https://github.com/security-force-monitor/nlp_starter_dataset}}

Annotating data is a way to describe what relevant information can be found inside a particular document, and how these bits of information are connected together. Annotations are used to train and assess the performance of NLP systems that are subsequently developed to perform the desired task.  SFM made the annotations using the Berkeley Rapid Annotation Tool (BRAT) \cite{brat_vis}, which has a graphical interface that the annotator can use to highlight and connect different information contained in the text. The types of information that can be annotated are described in a BRAT configuration file:

\vspace*{3px}
\begin{lstlisting}[
    basicstyle=\fontsize{7}{5}\ttfamily %\tiny, or \small or \footnotesize etc.
]
[entities]
Person
Organization
Rank
Title
Role
[relations]
is_posted	Arg1:Person, Arg2:Organization
has_title	Arg1:Person, Arg2:Title
has_role	Arg1:Person, Arg2:Role
has_rank	Arg1:Person, Arg2:Rank
<OVERLAP>	Arg1:Role, Arg2:Rank, <OVL-TYPE>:<ANY>
<OVERLAP>	Arg1:Title, Arg2:Role, <OVL-TYPE>:<ANY>
\end{lstlisting}
\vspace*{3px}

The ``entities" section above shows us that an annotator can decide a particular word or extract describes a ``Person" or an ``Organization", as well as biographical information like a ``Title", ``Rank" or ``Role" the person may have in that organization. The ``relations" section shows how these building blocks can be connected to each other:  a ``Person" can be ``Posted" to an ``Organization", and a ``Person" can have a ``Role" and a ``Rank'' during the course of that posting.  

Each document in the annotated corpus has two corresponding files: the first stores the raw text, the second the annotations made to that text. A third file contains the document title, date, and other metadata and is not used in the research. The text and annotation files share the same file name (a 36-character UUID) while the suffixes are `.txt' and `.ann', respectively. 

Annotations follow the BRAT Standoff format \cite{brat_standoff_format}. Named entities are identified by text-bounds, which use two numbers to locate the first character and the last character of a name entity. Relations are identified with two arguments, which are the ID's of the two name entities involved in the relation. Relations are directed. The following is an example of the annotations for the sentence mentioned above:

\vspace*{3px}
\begin{lstlisting}[
    basicstyle=\fontsize{7}{5}\ttfamily %\tiny, or \small or \footnotesize etc.
]
T1	Title_Role 35 38	GOC
T2	Title_Role 52 70	Officer Commanding
T3	Organization 71 90	3 Armoured Division
T4	Organization 98 111	Nigerian Army
R1 has_rank Arg1:T1 Arg2:T2	
R2 is_posted Arg1:T1 Arg2:T3	
R3 has_title Arg1:T1 Arg2:T4	
\end{lstlisting}
\vspace*{3px}

The  annotations can be also be visualized in the BRAT tool, an example of which is shown in Figure \ref{fig:brat_vis_screenshot}.

Our research task is to automate the extraction of such annotations from a raw-text input, and gain insight into way in which this capability could replace, substantially augment or otherwise assist researchers in performing this task.   The experimental system is not required to reconcile entities across documents or with an external dataset (entity linking). Even though our pilot system is basic, our aim is to quantify the system's performance and identify the key factors that affect this performance, sufficiently to say whether or not the task can be accomplished in a way that would make it worth implementing as part of SFM's research workflow. The following section discusses some of these factors. 

\section{Pilot System} 
\label{section:pilotstudy}

Our pilot system\footnote{Available at \url{https://github.com/security-force-monitor/sfm-graph-extractor}} addresses only two sub-tasks of the full knowledge graph extraction task: Named Entity recognition, and relation extraction. We aim to address entity linking in a future working paper. 

\subsection{Named Entity Recognition (NER)}
To extract name entities, we use BiLSTM-CNN-CRF model \cite{ma2016endtoend} in the traditional inside–outside–beginning (IOB) tagging framework \cite{Sang:1999:RTC:977035.977059}. Here is an example of the IOB tagging for the sentence in Figure \ref{fig:brat_vis_screenshot}:

\vspace*{3px}

\begin{lstlisting}[
    basicstyle=\fontsize{7}{5}\ttfamily %\tiny, or \small or \footnotesize etc.
]
General Officer Commanding 3     Armoured Division 
B-TTL   I-TTL   I-TTL      B-ORG I-ORG    I-ORG    

of    the   Nigerian Army  , Major General Jack  
I-ORG I-ORG I-ORG    I-ORG   B-RNK I-RNK   B-PER 

Nwaogbo, has again re-assured Nigerians that the 
I-PER    O   O     O          O         O    O   

Boko Haram insurgency would soon be contained.
O    O     O          O     O    O  O   
 
 
\end{lstlisting}

\vspace*{3px}

We first compute input representations for each token by applying a convolutional neural network (CNN) to compute character-level representations. CNNs have been shown to be effective at extracting morphological information \cite{DosSantos:2014:LCR:3044805.3045095, chiu-nichols-2016-named}. The output of the CNN layer is concatenated with pre-trained GloVe word embedding \cite{pennington-etal-2014-glove} to represent each token.\\
\indent Next, we compute context representations from the word-level representations by encoding the context using a BiLSTM \cite{Hochreiter:1997:LSM:1246443.1246450, pascanu2012difficulty}. Because we are using the IOB tagging format \cite{Sang:1999:RTC:977035.977059}, the label sequence follows certain rules. For instance, I-ORG cannot follow I-PER. Therefore, label sequences are modeled jointly using a conditional random field (CRF) \cite{Lafferty:2001:CRF:645530.655813}.\\
\indent 
Because the data provided by SFM for this task is relatively small \cite{SFM_dataset}, and part of it is needed for testing, we retrained the model on a combination of this data and a part of the CoNLL-2003 dataset \cite{tjong-kim-sang-de-meulder-2003-introduction}. The CoNLL-2003 dataset has two classes which also appears in our dataset: `Person' and `Organization'. In addition, we added a list of known organizations to our dataset. Since it is hard to draw a clear line between the class `Title' and the class `Role', which were specified as distinct in SFM's knowledge graph, we decided to collapse them into one class. 

The performance of the named entity model is shown in Table \ref{table:NER_eval}.

\begin{table*}[htp]
\centering
\begin{tabular}{||c c c c c c c||} 
 \hline
Class & \thead{True \\ Positives} & \thead{False \\ Positives} & \thead{False \\ Negatives} & Precision & Recall & F1 Score \\ 
 \hline\hline

Person & 87 & 13 & 6 & 0.87 & 0.94 & 0.90 \\ 
 \hline
Rank & 80 & 14 & 11 & 0.85 & 0.88 & 0.86 \\ 
 \hline
Organization & 103 & 33 & 31 & 0.76 & 0.77 & 0.76\\ 
 \hline
Title/Role & 85 & 20 & 23 & 0.81 & 0.79 & 0.80\\ 
 \hline
All Classes & 355 & 80 & 71 & 0.82 & 0.83 & 0.82\\ 
 \hline
\end{tabular}
\caption{NER model evaluation}
\label{table:NER_eval}
\end{table*}

\begin{table*}[htp]
\centering
\begin{tabular}{||c c c c c c c||} 
 \hline
 Method  & \thead{True \\ Positives} & \thead{False \\ Positives} & \thead{False \\ Negatives} & Precision & Recall & F1 score \\
 \hline
 \hline
 Nearest Person (Baseline) & 993 & 759 & 423 & 0.567 & 0.701 & 0.627\\
 Shortest Dep. Path (No constraint)  & 1083 & 651 & 333 & 0.625 & 0.765 & 0.687\\
 Shortest Dep. Path (With constraint)  & \textbf{1180*} & 559 & \textbf{236*} & 0.679 & \textbf{0.833*} & \textbf{0.748*}\\
 Neural Network (No constraint) & 1086 & 667 & 330 & 0.620 & 0.767 & 0.685\\
 Neural Network (With constraint) & 1103 & \textbf{450*} & 313 & \textbf{0.710*} & 0.779 & 0.743\\
 \hline
\end{tabular}
\caption{RE algorithms evaluation}
\label{table:RE_eval}
\end{table*}

\begin{table*}[htp]
\centering
\begin{tabular}{||c c c||} 
 \hline
 Component  & Time (Seconds) & Model Size (Parameters)\\
 \hline
 \hline
 NER & 1.54 & 6153100\\
 Dep. Parsing & 0.70 & 8791858\\
 Shortest Dep. Path & 0.0039 & N/A\\
 Neural Network & 0.051 & 294\\
 \hline
\end{tabular}
\caption{Average Processing Time per line \& Model sizes}
\label{table:pipeline_speed}
\end{table*}

\subsection{Relation Extraction}
In our pilot system, we experiment with three approaches to relation extraction: nearest person, shortest dependency path, and a neural network based approach. These approaches all share the same underlying idea: starting at each non-Person named entity $e$, our system tries to identify an entity of type Person in the same sentence that stands in a relation with $e$, and then tries to predict the type of this relation. We will look at each approach in turn.

\subsubsection{Nearest Person}
This baseline algorithm is based on the simple idea that a non-Person named entity is often related to a Person named entity nearby. For example, one could say ``General Lamidi Adeosun'' where `'`Lamidi Adeosun'' has the rank ``General'', as in one of the documents in our corpus. The algorithm, therefore, merely relates a non-Person named entity to the Person named entity immediately to the right; if there is no Person named entity to its right, then the algorithm relates it to the nearest Person entity no matter which side the Person entity is on. Although we did not expect performance of this system to be competitive, an immediate advantage of this simple technique is that its decisions are transparent.

\subsubsection{Shortest Dependency Path}
\label{section:sdp}
Instead of using the distances between named entities in raw text, we can take syntactic information into account. This method relates a non-Person named entity to the person named entity to which the dependency path is shortest. This relies largely on how well the dependency parser performs, so we used a state-of-the-art dependency parser \cite{nguyen-verspoor-2018-improved}. Since one named entity could span multiple tokens, we only use the shortest path among the various possible paths between the tokens of two named entities.

We find that constraining the algorithm to only choose between the two Persons that appear immediately to the left and to the right increases performance, at least on our data set. We use this constraint in the final version of our system. 

Assuming reasonable performance by the dependency parser, decisions of this heuristic dependency based approach are easy to trace.

\subsubsection{Neural Network}
In this approach, we use machine learning to predict relations based on dependency paths and named entity types. We use the phrase ``path pattern'' to refer to the list of edge types along a dependency path. The path patterns are encoded into one-hot vectors. Uncommon path patterns are treated as a single ``unknown'' category. Since there are multiple possible persons in a sentence, there will be a one-hot vector for each Person. We pass these multiple vectors as input to the network, so that it makes a joint decision over multiple candidate entities of type Person.

 In addition, the path length is concatenated to the one-hot vector to compensate for the loss of information when we replace the less frequent path patterns. Multiple one-hot + length vectors of different persons are concatenated together along with a small one-hot vector which encodes the type of the non-person name entity, which makes up the input of the neural network. 

\begin{figure}[H]
\centering
\includegraphics[width=8cm]{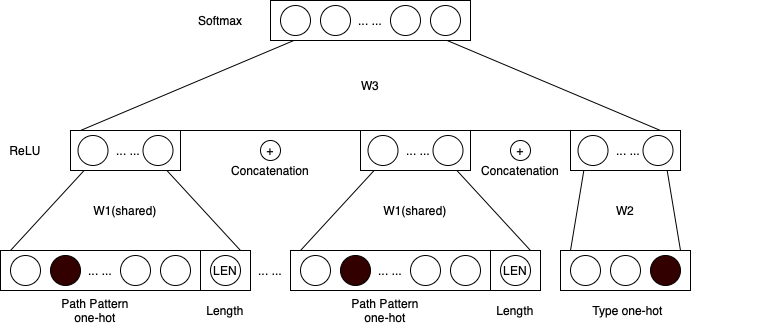}
\caption{The architecture of the neural network: The weight matrix $W_1$ for path pattern information is shared across all path pattern one-hot vectors, but it is distinct from the weight matrix $W_2$ for type one-hot vector. The softmax-activated output vector represents the probability of potential Person name entities.}
\label{fig:model}
\end{figure}

\indent {\bf Network architecture:} The first layer has a set of shared weights for those one-hot + length vectors and a separate set of weights for the name entity type vector. The second layer is a dense layer whose output is activated by a softmax layer. The softmax layer outputs a vector where the largest element corresponds to the target Person. The architecture is shown in Figure \ref{fig:model} and its performance is shown in Table \ref{table:RE_eval}.\\

The number of persons that the model could process within a sentence is limited to 7. If there are more than 7 persons in a sentence, we set the target to an all-zero vector. If a prediction does not correspond to any person, such as when there are only 3 persons in a sentence while the model predicts the fourth person, then we do not build any relations for the name entity.

To include the constraint mentioned in Section \ref{section:sdp}, we made the target a 3-element vector. If the first element is the largest, then the Person on the left side is the predicted person; if the second element is the largest, then the Person on the right side is the predicted Person; if the third element is the largest, then the model predicts that the correct Person should be some Person other than the two nearest Persons. When the model predicts that the Person is not nearby, we select the Person that has shortest dependency path excluding the two nearest Persons.

\begin{figure*}[htp]
\begin{subfigure}{\textwidth}
  \centering
  \includegraphics[width=15cm, height=3cm]{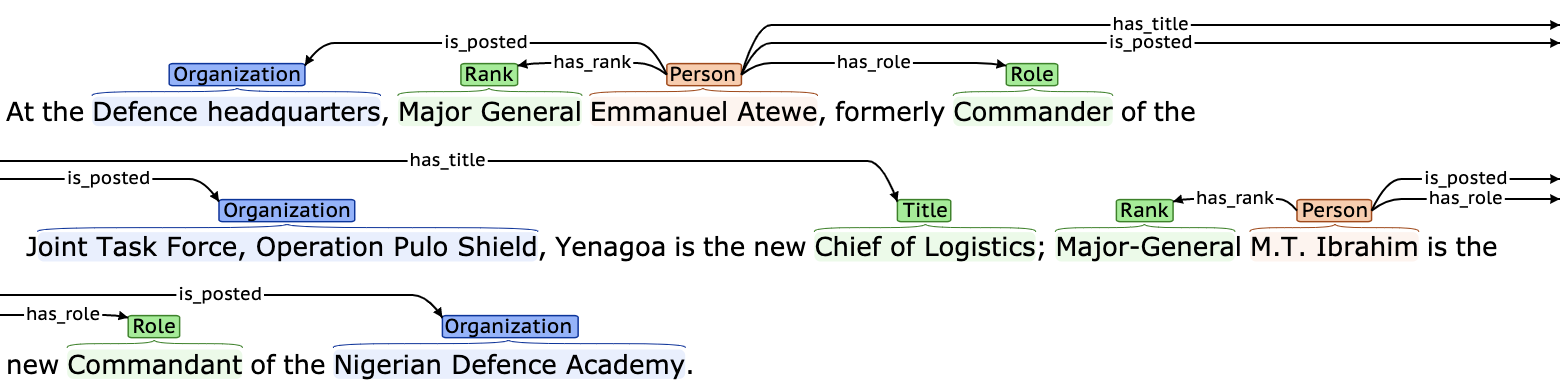}  
  \caption{True Annotations}
  \label{fig:ex1_true}
\end{subfigure}
\begin{subfigure}{\textwidth}
  \centering
  \includegraphics[width=15cm, height=3cm]{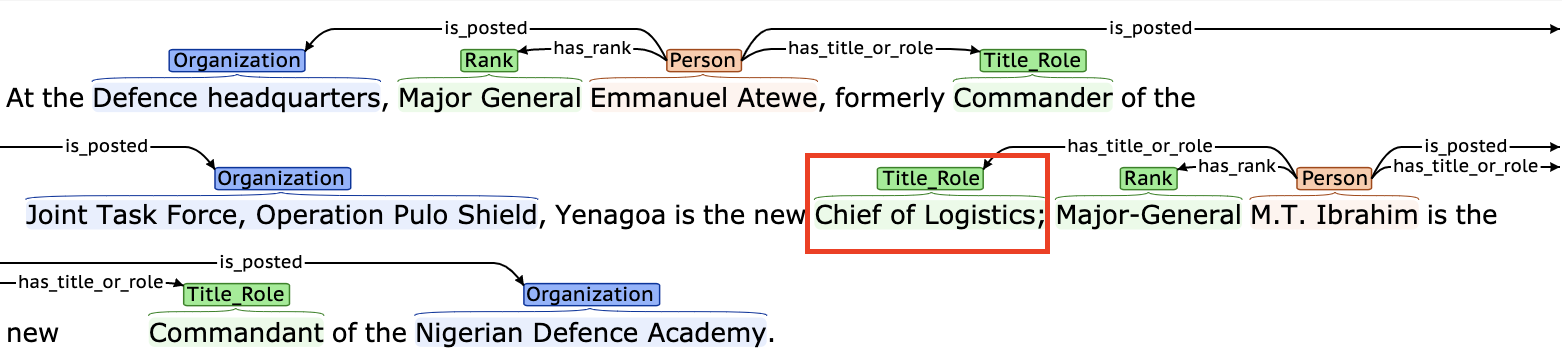}  
  \caption{Shortest Dependency Path}
  \label{fig:ex1_dep}
\end{subfigure}
\begin{subfigure}{\textwidth}
  \centering
  \includegraphics[width=15cm, height=3.5cm]{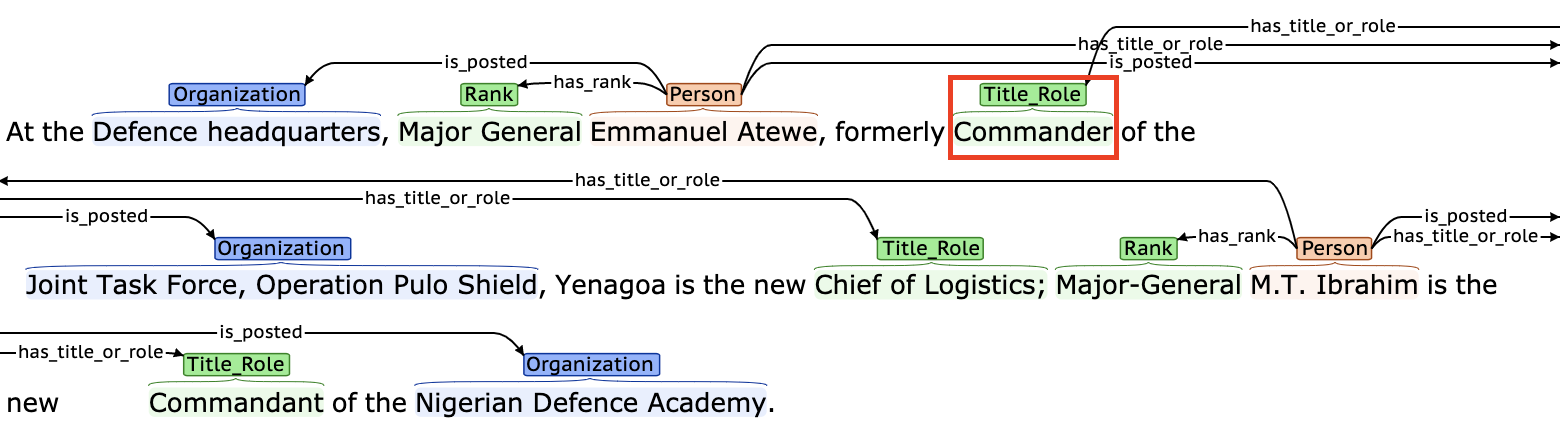}  
  \caption{Neural Network}
  \label{fig:ex1_nn}
\end{subfigure}
\caption{Two methods make different mistakes}
\label{fig:results_ex1}
\end{figure*}

\begin{figure*}[htp]
\begin{subfigure}{\textwidth}
  \centering
  \includegraphics[width=15cm, height=2cm]{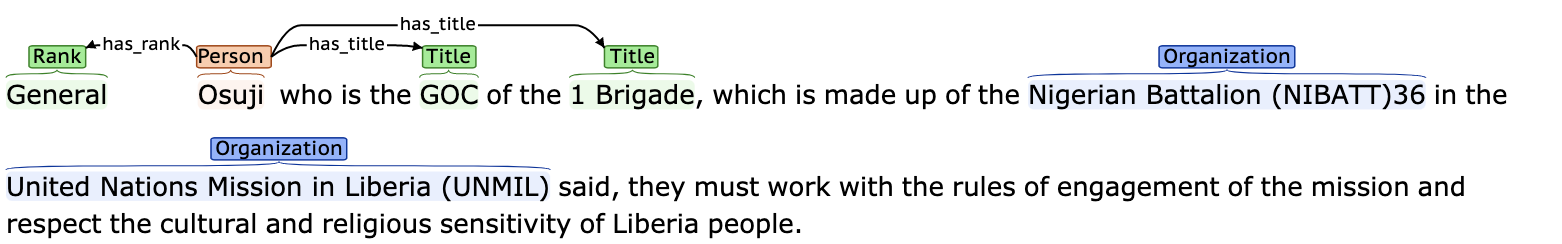}  
  \caption{True Annotations}
  \label{fig:ex2_true}
\end{subfigure}
\begin{subfigure}{\textwidth}
  \centering
  \includegraphics[width=15cm, height=2.8cm]{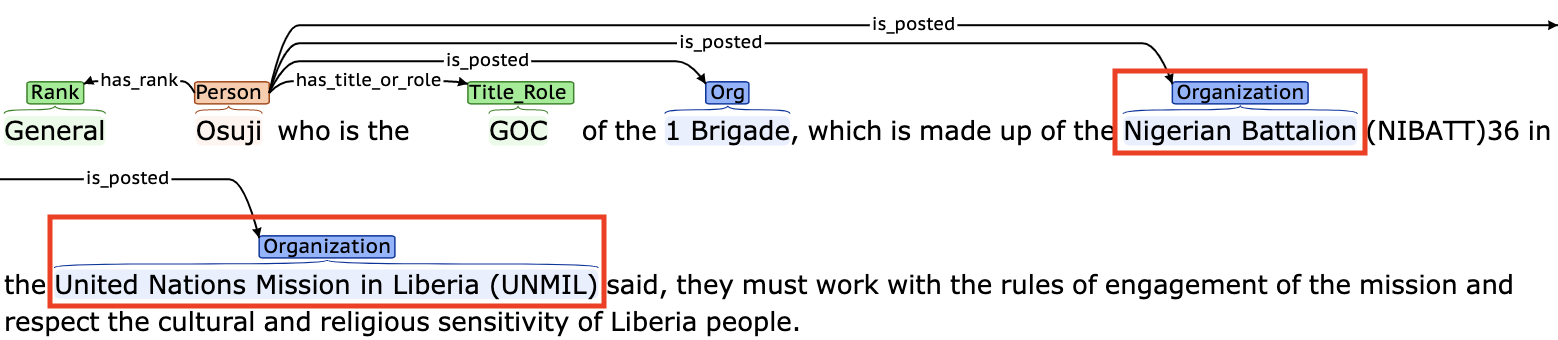}  
  \caption{Shortest Dependency Path}
  \label{fig:ex2_dep}
\end{subfigure}
\begin{subfigure}{\textwidth}
  \centering
  \includegraphics[width=15cm, height=1.8cm]{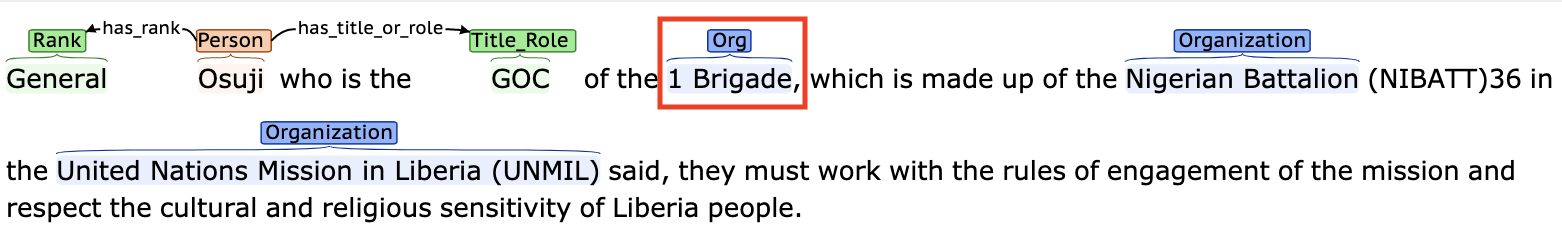}  
  \caption{Neural Network}
  \label{fig:ex2_nn}
\end{subfigure}
\caption{Neural Network is more conservative}
\label{fig:results_ex2}
\end{figure*}

\section{Experimental Results}
\label{section:expresults}
For sentences with relatively few named entities (for example, a sentence has only 1 Person named entity), both Shortest Dependency Path and Neural Network perform very well since there are not many choices to make. When there are more named entities in a sentence, it becomes harder to make a correct choice and that is where the two methods make different predictions.

The neural network method looks at the path length, its pattern and the named entity type, so it sees more information than the shortest dependency path method, which only looks at path length. These two additional pieces of information sometimes do help make a better prediction, but they can confuse the neural network as well. In Figure \ref{fig:results_ex1}, the shortest dependency path method falsely related the name entity ``Chief of Logistics'' to the person ``M. T. Ibrahim''. ``Chief of Logistics'' is closer to ``M. T. Ibrahim'' than to ``Emmanuel Atewe'' in the dependency tree, but the additional information helped the neural network method make the correct choice. But when predicting ``Commander'', simply using the length of the dependency path yields the right relation.

The neural network method has fewer false positives, since the shortest dependency path method is forced to build a relation for every non-Person name entity while the neural network has the option not to build a relation. Therefore, the neural network has fewer false positives and more false negatives. This indicates that the neural network is slightly too conservative when deciding whether there exists a correct relation for a given non-Person name entity. One example is shown in Figure \ref{fig:results_ex2}.

There are some obvious limitations of our algorithms. First, all three algorithms only relate two named entities when they are in the same sentence. When there are relations that cross sentences, our algorithms will not be able to capture them. Second, sometimes, one named entity can be related to multiple other entities. Our algorithms only relate one non-Person name entity to a single Person name entity. Third, we have not implemented the functionality to collect and reconcile the information of Person named entities that exist in different documents.

We also measured the average processing time of different components in the pipeline, as shown in Table \ref{table:pipeline_speed}. The results are based on three independent but identical measurements of the processing time on our entire dataset. The numbers are obtained by averaging the measured time after discarding obvious outliers. The measurements are conducted on a MacBook Pro with a quad-core CPU @ 4.1GHz Max. It is worth noting that the processing time of NER and dependency parsing is dependent on the length of the sentences and the processing time of the shortest dependency parsing algorithm or the neural network is dependent on the amount of named entities in a line.

\section{Discussion and conclusion}
\label{section:conclusion}

The results show that an NLP system can perform the research task with a \textit{fair to good} degree of accuracy, albeit with some clear limitations. The system skews a little towards recall, which is preferable where a subsequent human review is intended. Demonstrating this capability meets the first objective of this collaboration. However, the results do not tell us whether its performance is tolerable within SFM's research workflow (and by extension that of other human rights researchers). Understanding this requires implementing the system in a way that enables SFM to accept, reject or quickly update the proposals it makes, and assessing whether this creates savings in time and resource as compared to doing the work wholly "by hand". Subsequent work will focus on this next, implementation step. 

Although the present paper is mostly technical and focused on practical application,  throughout the authors' collaboration we have ranged across the wider matters of NLP and the challenges of technology implementation within the human rights domain. The potential that NLP represents is set against the sector's considerable financial and technical capacity constraints and a dearth in transparent examples of successful NLP use within it. Surrounding this are concerns about the human rights implications of NLP methods themselves:  the discriminatory potential of the datasets used to train them; the dominance of government and corporate actors in their technical development; and, implementations that infringe human rights directly. In future papers drawn from our research, we aim to assess how these affect the desirability and feasibility of NLP use within the wider non-profit domain.

\end{multicols}
\printbibliography

@misc{ma2016endtoend,
    title={End-to-end Sequence Labeling via Bi-directional LSTM-CNNs-CRF},
    author={Xuezhe Ma and Eduard Hovy},
    year={2016},
    eprint={1603.01354},
    archivePrefix={arXiv},
    primaryClass={cs.LG}
}

@inproceedings{DosSantos:2014:LCR:3044805.3045095,
     author = {Dos Santos, Cícero Nogueira and Zadrozny, Bianca},
     title = {Learning Character-level Representations for Part-of-speech Tagging},
     booktitle = {Proceedings of the 31st International Conference on International Conference on Machine Learning - Volume 32},
     series = {ICML'14},
     year = {2014},
     location = {Beijing, China},
     pages = {II-1818--II-1826},
     url = {http://dl.acm.org/citation.cfm?id=3044805.3045095},
     acmid = {3045095},
     publisher = {JMLR.org},
}

@article{chiu-nichols-2016-named,
    title = "Named Entity Recognition with Bidirectional {LSTM}-{CNN}s",
    author = "Chiu, Jason P.C.  and
     Nichols, Eric",
    journal = "Transactions of the Association for Computational Linguistics",
    volume = "4",
    year = "2016",
    url = "https://www.aclweb.org/anthology/Q16-1026",
    doi = "10.1162/tacl_a_00104",
    pages = "357--370"
}

@misc{pascanu2012difficulty,
    title={On the difficulty of training Recurrent Neural Networks},
    author={Razvan Pascanu and Tomas Mikolov and Yoshua Bengio},
    year={2012},
    eprint={1211.5063},
    archivePrefix={arXiv},
    primaryClass={cs.LG}
}

@article{Hochreiter:1997:LSM:1246443.1246450,
    author = {Hochreiter, Sepp and Schmidhuber, J\"{u}rgen},
    title = {Long Short-Term Memory},
    journal = {Neural Comput.},
    issue_date = {November 15, 1997},
    volume = {9},
    number = {8},
    month = nov,
    year = {1997},
    issn = {0899-7667},
    pages = {1735--1780},
    numpages = {46},
    url = {http://dx.doi.org/10.1162/neco.1997.9.8.1735},
    doi = {10.1162/neco.1997.9.8.1735},
    acmid = {1246450},
    publisher = {MIT Press},
    address = {Cambridge, MA, USA},
}

@inproceedings{Sang:1999:RTC:977035.977059,
     author = {Sang, Erik F. Tjong Kim and Veenstra, Jorn},
     title = {Representing Text Chunks},
     booktitle = {Proceedings of the Ninth Conference on European Chapter of the Association for Computational Linguistics},
     series = {EACL '99},
     year = {1999},
     location = {Bergen, Norway},
     pages = {173--179},
     numpages = {7},
     url = {https://doi.org/10.3115/977035.977059},
     doi = {10.3115/977035.977059},
     acmid = {977059},
     publisher = {Association for Computational Linguistics},
     address = {Stroudsburg, PA, USA},
}

@inproceedings{Lafferty:2001:CRF:645530.655813,
     author = {Lafferty, John D. and McCallum, Andrew and Pereira, Fernando C. N.},
     title = {Conditional Random Fields: Probabilistic Models for Segmenting and Labeling Sequence Data},
     booktitle = {Proceedings of the Eighteenth International Conference on Machine Learning},
     series = {ICML '01},
     year = {2001},
     isbn = {1-55860-778-1},
     pages = {282--289},
     numpages = {8},
     url = {http://dl.acm.org/citation.cfm?id=645530.655813},
     acmid = {655813},
     publisher = {Morgan Kaufmann Publishers Inc.},
     address = {San Francisco, CA, USA},
}

@inproceedings{pennington-etal-2014-glove,
    title = "{G}love: Global Vectors for Word Representation",
    author = "Pennington, Jeffrey  and
      Socher, Richard  and
      Manning, Christopher",
    booktitle = "Proceedings of the 2014 Conference on Empirical Methods in Natural Language Processing ({EMNLP})",
    month = oct,
    year = "2014",
    address = "Doha, Qatar",
    publisher = "Association for Computational Linguistics",
    url = "https://www.aclweb.org/anthology/D14-1162",
    doi = "10.3115/v1/D14-1162",
    pages = "1532--1543",
}

@inproceedings{tjong-kim-sang-de-meulder-2003-introduction,
    title = "Introduction to the {C}o{NLL}-2003 Shared Task: Language-Independent Named Entity Recognition",
    author = "Tjong Kim Sang, Erik F.  and
      De Meulder, Fien",
    booktitle = "Proceedings of the Seventh Conference on Natural Language Learning at {HLT}-{NAACL} 2003",
    year = "2003",
    url = "https://www.aclweb.org/anthology/W03-0419",
    pages = "142--147",
}

@inproceedings{nguyen-verspoor-2018-improved,
    title = "An Improved Neural Network Model for Joint {POS} Tagging and Dependency Parsing",
    author = "Nguyen, Dat Quoc  and
      Verspoor, Karin",
    booktitle = "Proceedings of the {C}o{NLL} 2018 Shared Task: Multilingual Parsing from Raw Text to Universal Dependencies",
    month = oct,
    year = "2018",
    address = "Brussels, Belgium",
    publisher = "Association for Computational Linguistics",
    url = "https://www.aclweb.org/anthology/K18-2008",
    doi = "10.18653/v1/K18-2008",
    pages = "81--91"
}

@article{nyt2018,
    author = {Searcey, D. and Akinwotu E.},
    title = {Nigeria Says Soldiers Who Killed Marchers Were Provoked. Video Shows Otherwise.},
    year = {2018},
    month = {12},
    day = {17},
    url = {https://www.nytimes.com/2018/12/17/world/africa/nigeria-military-abuses.html},
    journal = {New York Times},
}

@misc{wilson_2017, 
    howpublished =  "\url{https://securityforcemonitor.org/2017/04/11/why-i-started-the-security-force-monitor/}",
    title = {Why I Started The Security Force Monitor},
    author={Wilson, T.}, 
    year={2017}, 
    month={4}
}

@misc{philippines2019,  
    title = {Investigating Drug Related Killings in the Philippines},
    year={2019},
    howpublished = "\url{https://securityforcemonitor.org/2019/08/19/investigating-drug-related-killings-in-the-philippines/"}",
    author={Security Force Monitor},
}

@misc{mexico2018,     
    title = {Prosecutor of the International Criminal Court receives complaint of crimes against humanity by the Mexican Army},
    year = {2018},
    howpublished = {\url{https://securityforcemonitor.org/2018/06/11/prosecutor-of-the-international-criminal-court-receives-complaint-of-crimes-against-humanity-by-the-mexican-army/}},
    author={Security Force Monitor},
}

@misc{myanmar2018,     
    title = {The structure and operations of the Myanmar Army in Rakhine State: a review of open source evidence},
    year = {2018},
    howpublished = {\url{https://securityforcemonitor.org/2018/11/20/myanmar-army-in-rakhine-state-structure-and-operations/}},
    author={Security Force Monitor},
}

@misc{vanguard2012,
    author = {Taye Obateru},
    title = {Boko Haram ‘ll soon be contained – GOC},
    year = {2012},
    month = {7},
    day = {2},
    howpublished = {\url{http://www.vanguardngr.com/2012/07/boko-haram-ll-soon-be-contained-goc/}}
}

@manual{brat_standoff_format,
    title  = "BRAT Standoff Format",
    url    = "https://brat.nlplab.org/standoff.html",
    organization = "brat",
    year = "2020",
}

@manual{brat_vis,
    title  = "BRAT Rapid Annotation Tool",
    url    = "https://brat.nlplab.org/index.html",
    organization = "brat",
    year = "2020",
}

@misc{SFM_dataset,
    title = {WhoWasInCommand},
    note  = {https://whowasincommand.com/},
    author={Security Force Monitor},
    year = {2020},
}
\end{document}